\documentclass[letterpaper]{article} 
\usepackage{aaai2026}  
\usepackage{times}  
\usepackage{helvet}  
\usepackage{courier}  
\usepackage[hyphens]{url}  
\usepackage{graphicx} 
\usepackage{natbib}  
\usepackage{caption} 
\usepackage{algorithm}
\usepackage{algorithmic}

\usepackage[table]{xcolor} 
\newcommand{\subsubsec}[1]{%
    \vspace{1em}
    \noindent%
    \textbf{\fontsize{10}{12}\selectfont #1.}%
    \quad
}
\PassOptionsToPackage{table}{xcolor}
\usepackage{booktabs}    
\usepackage{multirow}    
\usepackage{makecell}    
\usepackage{array}       
\usepackage{graphicx}  
\usepackage{float}
\usepackage{soul}
\usepackage{booktabs}
\usepackage{algorithm}
\usepackage{algorithmic}
\usepackage{amsmath}
\usepackage{amssymb}
\usepackage{mathrsfs}
\usepackage{graphicx}

\newtheorem{definition}{Definition}
\newtheorem{theorem}{Theorem}
\usepackage{caption} 
\usepackage{subcaption}
\usepackage{newfloat}
\usepackage{listings}
\DeclareCaptionStyle{ruled}{labelfont=normalfont,labelsep=colon,strut=off} 
\floatstyle{ruled}
\newfloat{listing}{tb}{lst}{}
\floatname{listing}{Listing}
\title{Learning Optimal Prompt Ensemble for Multi-source Visual Prompt Transfer}

\author{
Enming Zhang\textsuperscript{1} \quad 
Liwen Cao\textsuperscript{2} \quad 
Yanru Wu\textsuperscript{1} \quad 
Zijie Zhao\textsuperscript{1} \quad 
Yang Li\textsuperscript{1}\thanks{Corresponding author. Email: yangli@sz.tsinghua.edu.cn}
\\
\textsuperscript{1}Tsinghua Shenzhen International Graduate School, Tsinghua University \thanks{Shenzhen Key Laboratory of Ubiquitous Data Enabling} \\
\textsuperscript{2}Southeast University \quad 
}
\date{}  

\begin{document}
\copyrighttext{Copyright \textcopyright{} 2024, Association for the Advancement of Artificial Intelligence (www.aaai.org). All rights reserved.}

\maketitle

\begin{abstract}
Prompt tuning has emerged as a lightweight strategy for adapting foundation models to downstream tasks, particularly for resource-constrained systems. As pre-trained prompts become valuable assets, combining multiple source prompts offers a promising approach to enhance generalization for new tasks by leveraging complementary knowledge. However, naive aggregation often overlooks different source prompts have different contribution potential to the target task.
To address this, we propose HGPrompt, a dynamic framework that learns optimal ensemble weights. These weights are optimized by jointly maximizing an information-theoretic metric for transferability and minimizing gradient conflicts via a novel regularization strategy. Specifically, we propose a differentiable prompt transferability metric to captures the discriminability of prompt-induced features on the target task. Meanwhile, HGPrompt match the gradient variances with respect to different source prompts based on Hessian and Fisher Information, ensuring stable and coherent knowledge transfer while suppressing gradient conflicts among them.
Extensive experiments on the large-scale VTAB benchmark demonstrate the state-of-the-art performance of HGPrompt, validating its effectiveness in learning an optimal ensemble for effective multi-source prompt transfer.
\end{abstract}

\section{Introduction}
\begin{figure}[t!]
  \centering
  \includegraphics[width=0.47\textwidth]{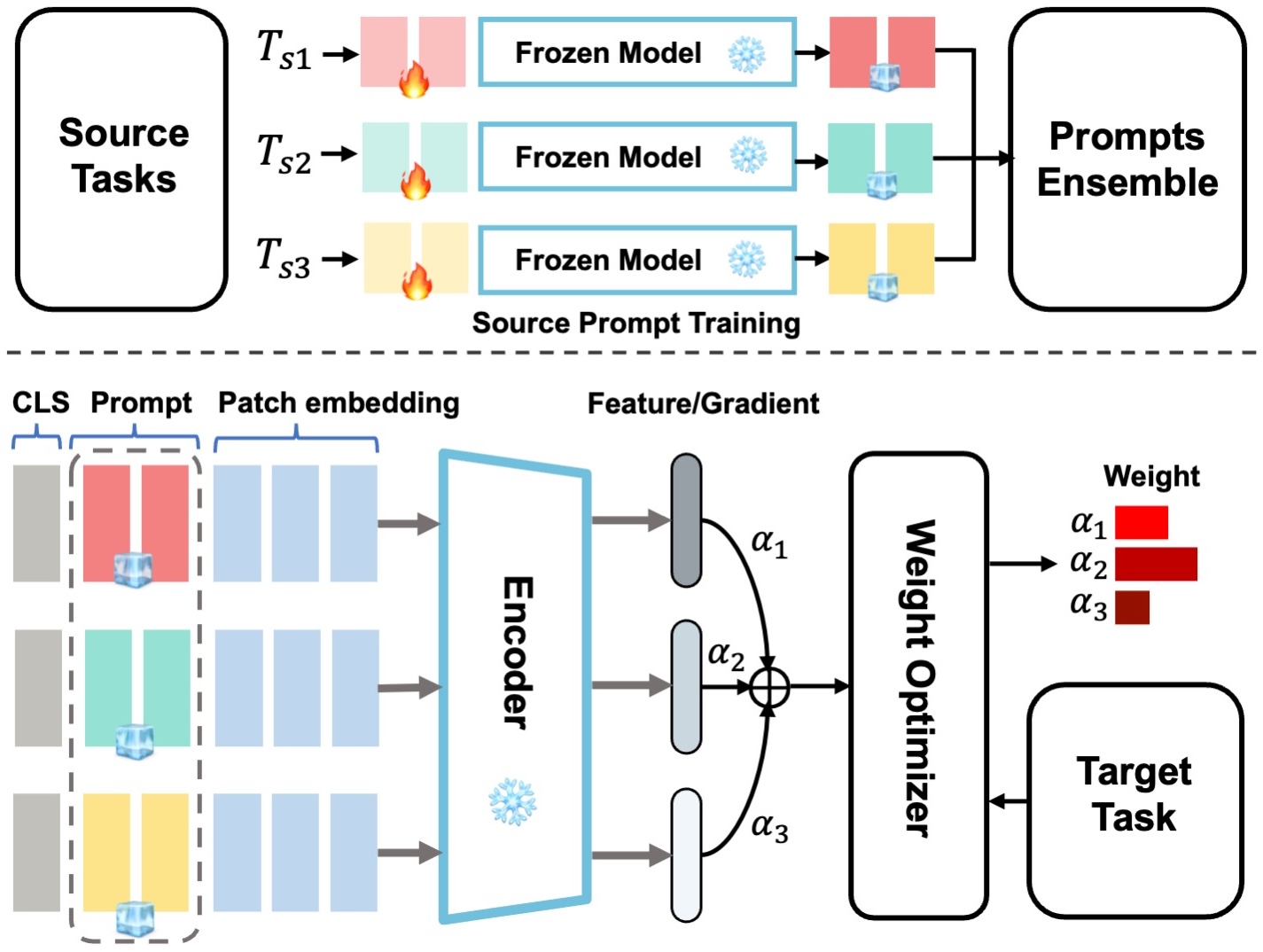} 
  \caption{Multi-source prompt transfer framework. Task-specific prompts are tuned via a frozen backbone and statically aggregated for target initialization. Our approach dynamically optimizes source weights through single forward-backward propagation, learning prompt aggregation via an optimization module.}
  \label{fig:first}
  \vspace{-0.3cm}
\end{figure}
With the development of expanding datasets, novel architectures, and improved training algorithms~\cite{chen2020simple}, a significant number of vision foundation models have been developed \cite{radford2021learning,dosovitskiy2020image,liu2021swin}.
Transformer-based pre-trained vision models (PVMs) demonstrate exceptional efficacy across diverse tasks, including image classification and semantic segmentation. While these models exhibit impressive capability, adapting them to downstream applications still presents notable challenges. Full model fine-tuning becomes impractical given the substantial parameter volumes and challenges in low-data scenarios. This paradigm shift has made prompt tuning \cite{huang2022learning,lester-etal-2021-power, zhou2022learning} a key adaptation strategy. By freezing PVMs and adding learnable prompt tokens, it achieves competitive performance with only 0.4\% parameter updates, significantly fewer than full fine-tuning.

The increasing sophistication of prompt learning has established well-generalized prompts as valuable intellectual assets \cite{schick-schutze-2021-exploiting}. This evolution has fostered a practical ecosystem where users can access provider task-specific prompts while maintaining data privacy and model integrity. With the availability of multiple prompts from the prompt pool, these prompts can be utilized in an ensemble way by concurrently assembling them and transferring them to a single pre-trained model, as illustrated in Fig.~\ref{fig:first} ~\cite{sanh2021multitask, wang2023multitask}. However, simply concatenating or averaging source prompts often proves suboptimal, as the knowledge encoded in different prompts may contribute unevenly to the target task and can even lead to representation collapse \cite{standley2020tasks}. 

Effective multi-source transfer necessitates the assignment of adaptive weights to each source prompt. However, conventional methods \cite{vu2021spot, asai-etal-2022-attempt,su-etal-2022-transferability, panda-2024} predominantly evaluate the transferability of each prompt in isolation. This approach fails to account for potential interdependencies when prompts are combined within an ensemble, overlooking complementary effects that can significantly alter overall transferability. Furthermore, existing techniques often rely on heuristic methods, such as computing similarity between prompts parameters \cite{vu2021spot}, which typically lack a rigorous theoretical foundation.

To overcome these limitations, we propose a lightweight and theoretically reliable framework that dynamically learns optimal prompt weights. Distinct from prior methods evaluating each prompt in isolation, our key innovation lies in evaluating the transferability of feature ensemble induced by the aggregated prompts. Specifically, we learn the prompt weights by maximizing the H-score, a theoretically grounded, differentiable metric to quantify feature transferability. Unlike conventional approaches relying on heuristics, our method provides an explicit and interpretable measure of each prompt's contribution to the ensemble's transferability, rooted in information-theoretic and statistical principles \cite{xu2022information}.

Moreover, aggregating multiple prompts often introduces detrimental interference between their gradients, which leads to unstable optimization dynamics and suboptimal solution. Building upon the theoretical insight that similar Hessians and Fisher Information reduce inconsistencies in the loss landscape \cite{parascandolo2020learning, shi2021gradient, rame2022fishr} , we introduce a simple yet general gradient alignment regularization term in our optimization framework. Specifically, this term match gradient variance from the different source prompts. Minimizing this term encourages consensus during optimization. By resolving these inherent gradient conflicts, our approach develops a prompt ensemble with robust and consistent 

Our method achieves state-of-the-art performance through extensive evaluations on the large-scale VTAB benchmark \cite{zhai2019large}, consistently outperforming competitive strategies such as PANDA \cite{panda-2024}, SPoT, and ATTEMPT. Our approach establishes new benchmarks for future research on multi-source visual prompt transfer. The source code is available in the supplementary material.

\section{Related Work}
\subsection{Parameter-efficient Transfer Learning}

Parameter-efficient transfer learning is crucial for adapting large pre-trained models. In NLP, methods like adapters \cite{houlsby2019parameter}, BitFit \cite{ben-zaken-etal-2022-bitfit}, and LoRA \cite{hu2021lora} tune only 1-5\% of parameters. For vision, early work focused on ConvNets (e.g., residual adapters \cite{rebuffi2017learning}), but vision Transformers \cite{dosovitskiy2020image} introduced new challenges. While some NLP techniques (e.g., adapters \cite{chen2022adaptformer}) transfer directly, vision-specific approaches like VPT \cite{jia2022visual} (learnable tokens) and VP \cite{bahng2022exploring} (pixel-level perturbations) achieve high efficiency with minimal input-space modifications.

\begin{figure*}[t!]
  \centering
  \includegraphics[width=0.99\textwidth]{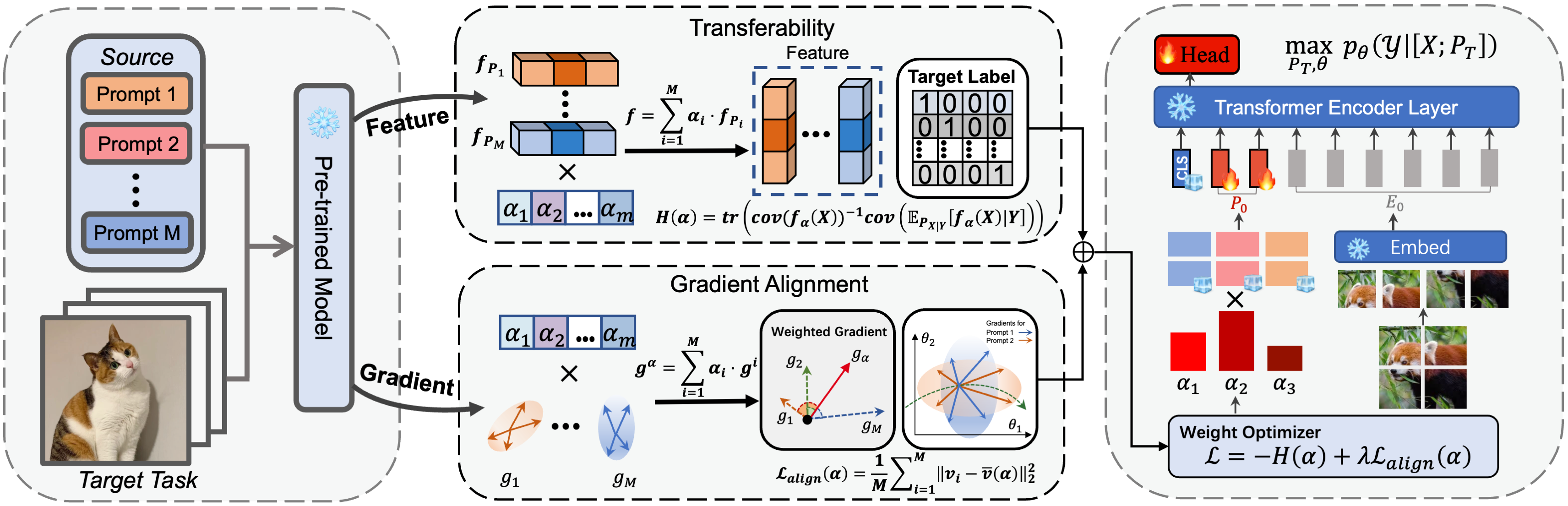} 
  \caption{Overview of the framework. Given an input image $X$, the system generates $M$ distinct feature representations $\{f_i\}_{i=1}^M$ and corresponding gradients $\{g^i\}_{i=1}^M$ through multiple source prompts. These features and gradients are fused using learnable weights $\boldsymbol{\alpha}$ to produce the final combined feature $f_{\boldsymbol{\alpha}}$ and gradient $g^{\boldsymbol{\alpha}}$. The Transferability term evaluates the fused feature distribution against the target class label, and the Gradient Alignment Regularization aligns the prompt gradient variance. The Weight Optimizer jointly optimizes these dual objectives to determine the optimal source weights $\boldsymbol{\alpha}$, which subsequently initialize the target prompt.}
  \label{fig:main}
  \vspace{-0.3cm}
    
\end{figure*} 

\subsection{Transferability Estimation}

Prompt transferability builds on task transferability research \citep{zamir2018taskonomy}, as prompts guide frozen models \cite{feng-2023-learning}. Existing metrics for task transferability \citep{ding2024model, tran2019transferability} can inform prompt evaluation. Information-theoretic approaches like H-score \citep{bao2019information, ibrahim2022newer, wu2024h}, LEEP \citep{nguyen2020leep, agostinelli2022transferability}, and LogME \citep{you2021logme} assess feature discriminability and performance prediction. Optimal transport methods like OTCE \citep{tan2021otce, tan2024transferability} also measure domain-task differences. These foundations support prompt transferability understanding.

\subsection{Multi-source Prompt Tuning}
Prompt-tuning on smaller pre-trained models often under-performs and is highly sensitive to prompt initialization, as evidenced by prior studies \cite{huang2022learning, lester-etal-2021-power}. To address these limitations, Prompt Transfer (PoT) methods have been proposed \cite{vu2021spot, su-etal-2022-transferability}, which leverage soft prompts learned on source tasks to initialize prompts for target tasks, thereby improving tuning efficiency and performance. SPOT \cite{vu2021spot} explored the use of metrics to predict the best source tasks for prompt transfer, and in parallel, \cite{su-etal-2022-transferability} emphasized how prompt-induced neuron activations play a crucial role in transferability. In addition to single-task transfer, PoT methods have been extended to multi-task settings. For example, ATTEMPT \cite{asai-etal-2022-attempt} proposed mechanisms to aggregate knowledge from multiple source tasks, using attention mechanisms strategies to initialize target prompts. PANDA \cite{panda-2024} explicitly addresses the issue of prior knowledge forgetting by distilling task-specific knowledge into the target prompt.

\section{Preliminary}
\subsection{Visual Prompt Tuning}
Visual Prompt Tuning (VPT) is a parameter-efficient transfer learning paradigm that adapts pre-trained vision transformers to downstream tasks by learning task-specific prompt embeddings while keeping the original model parameters frozen. This approach introduces a small set of learnable parameters in the form of prompt tokens, which are prepended to the input sequence, enabling efficient adaptation to new tasks without modifying the underlying model architecture. The key advantage of VPT lies in its ability to leverage the rich representations learned by large-scale pre-trained models while requiring significantly fewer trainable parameters compared to full fine-tuning.

Formally, given a pre-trained Transformer with embedding dimension $d$, we introduce $m$ learnable prompt tokens $P = [p_1, \ldots, p_m] \in \mathbb{R}^{m \times d}$. For an input image $X$ with patch embeddings $E(X) \in \mathbb{R}^{n \times d}$, the combined input sequence becomes $[P; E(X)] \in \mathbb{R}^{(m+n) \times d}$, where $m$ is the prompt length and $n$ is the number of image patches. The model parameters $\theta$ remain fixed during training, with gradients only propagating through the prompt embeddings $P$. The prediction probability for class $Y$ is given by:

\begin{equation}
\mathrm{Pr}_{\theta}(Y|X; P) = \frac{\exp(f_Y([P; E(X)]; \theta))}{\sum_{i=1}^C \exp(f_i([P; E(X)]; \theta))},
\end{equation}
where $C$ denotes the number of classes, and $f_i(\cdot)$ represents the pre-trained model's logit output for class $i$. This formulation allows the model to adapt to new tasks by learning task-specific context through the prompt tokens.

\subsection{Multi-Source Prompt Transfer}
In many real-world scenarios, we often have access to multiple source prompts that can be utilized for the target task. Multi-source prompt transfer aims to harness these related prompts to enhance performance on the target task. Given $\kappa$ source tasks $\mathcal{S} = \{S_i\}_{i=1}^\kappa$ along with their corresponding optimized prompts $\{P_i\}_{i=1}^\kappa$, our goal is to construct a target prompt $P_T$ for a new task $T$ by optimally combining the source prompts based on their relevance to the target task.

Let $M \leq \kappa$ denote the number of selected source prompts. We fix the hyperparameters $\boldsymbol{\alpha} = (\alpha_1, \ldots, \alpha_M)$ satisfying $\sum_{i=1}^M \alpha_i = 1$ and $\alpha_i \geq 0$.  Then we simultaneously optimizes both the header parameters $\theta$ and the target prompt $P_T$, where $P_T$ is initialized by a convex combination $P_T = \sum_{i=1}^M \alpha_i P_i$ of the frozen source prompts $\{P_i\}_{i=1}^M$.   
\begin{equation}
\max_{P_T, \theta}  \mathbb{E}_{(x,y) \sim \mathcal{D}_T} \left[\log P_{\theta}(y|[x; P_T])\right]
\end{equation}
This joint optimization learns both task-specific header parameters and target prompt formed by the weighted combination of source prompts. Crucially, the optimization landscape of $P_T$ is highly sensitive to this initialization, making the choice of $\boldsymbol{\alpha}$ a critical factor in final performance. This underscores why learning optimal combination weights is paramount. The weights $\alpha_i$ maintain interpretability by reflecting the relative importance of each source task to the target task. Our method mainly focuses on learning $\boldsymbol{\alpha}$, whose optimized values not only improve transfer performance but also can reveal task relationships and transferability insights.

\section{Methodology}
To dynamically learn optimal weights for source prompts, we propose a lightweight framework that jointly maximizes an information-theoretic transferability metric while matching gradient variance through a novel regularization strategy, as illustrated in Fig.~\ref{fig:main}. First, we present the mathematical formulation of the H-score based transferability metric and establish its theoretical reliability for optimization. Second, we provide a detailed explanation of the gradient alignment regularization, including its theoretical basis and intuition. The framework is designed to be both lightweight and interpretable, capable of serving as a plug-in module for multi-source prompt transfer scenarios.

\subsection{Measuring Prompt Ensemble Transferability}

To overcome the limitations of previous heuristic prompt ensemble strategies that treat prompts independently, we adopt a transferability metric to quantify each prompt's contribution to the combined ensemble. Specifically, we introduce an information theoretic metric for feature transferability based on H-score \cite{bao2019information,xu2022information}.
Unlike conventional assessments that assume transferability correlates with parameter similarity, our proposed metric focuses on the intrinsic informativeness of prompt-induced features, explicitly evaluating the effectiveness of prompt ensembles for the target task. The mathematical formulation of H-score is defined as follows:

\begin{definition}
With input data $x$, label $y$ and feature extractor $f(x)$ (a zero-mean feature function). The one-sided H-score of $f$ with regard to the task casting $x$ to $y$ is:
\begin{equation}
    H(\boldsymbol{f}) = \operatorname{tr}\left(\operatorname{cov}(f(X))^{-1}\operatorname{cov}\left(\mathbb{E}_{P_{X\mid Y}}[f(X)|Y]\right)\right).
\end{equation}

\end{definition}
The full derivation is provided in the Appendix. This formulation admits an intuitive interpretation:  
A high H-score indicates larger inter-class discriminability, characterized by  \(\operatorname{cov}\left( \mathbb{E}[f(X)|Y] \right)\), and minimized feature redundancy, reflected in a small \(\operatorname{tr}(\operatorname{cov}(f(X)))\).
Thus, elevated H-scores signify that prompt successfully elicits transfer-effective features from the model.

Given a frozen visual encoder $f_\theta$ and $M$ source prompts $\{P_i\}_{i=1}^M$ pre-trained on distinct tasks, for an input image $X \in \mathcal{X}$, the $i$-th source prompt feature extraction is defined as:
\begin{equation}
f_{P_i}(X) = f_\theta\left([x_0; P_i; E(X)]\right) \in \mathbb{R}^h
\label{eq:prompt_feature}
\end{equation}
where $E(X) \in \mathbb{R}^{n \times d}$ denotes image patch embeddings, $x_0 \in \mathbb{R}^d$ the [CLS] token, and $h$ the feature dimension. The optimal combination weights $\alpha$ are determined by maximizing the H-score of the weighted feature sum, which yields the most transferable prompted feature representation.
\begin{definition}
\label{def:optimal_weight}
Given source-specific features \(\{f_{P_j}\}_{j=1}^M\), the optimal feature weights \(\boldsymbol{\alpha} = (\alpha_1,\dots,\alpha_M)^\top \in \mathbb{R}^M\) are determined by:
\begin{equation}
\boldsymbol{\alpha}^* = \arg\max_{\boldsymbol{\alpha}} H\left( \sum\nolimits_{j=1}^M \alpha_j \cdot f_{P_j} \right) 
\quad \text{s.t.} \quad \sum\nolimits_{j=1}^M \alpha_j = 1
\end{equation}
\end{definition}
We then verify the benign property of the proposed optimization problem by proving that the optimal objective H-score is a convex function of $\boldsymbol{\alpha}$.

\begin{theorem}
\label{thm:h_convex}
Given input data $X$, labels $Y$, and  $\{f_{P_i}\}_{i=1}^n$ with $\sum_{i=1}^n \alpha_i = 1$, the H-score of the weighted feature is a convex quadratic form:
\begin{equation}
\begin{aligned}
H(f) = H\left( \sum\nolimits_{i=1}^n \alpha_i f_i \right) = 
\sum_{i=1}^n \sum_{j=1}^n \alpha_i \alpha_j \\ 
\cdot \operatorname{tr}\left( \mathbb{E}_{P_Y} \left[ \mathbb{E}_{P_{X|Y}}[f_i(X)|Y] \cdot \mathbb{E}_{P_{X|Y}}[f_j(X)|Y]^\top \right] \right)
\end{aligned}
\end{equation}
\end{theorem}
With above theoretical guarantee, the optimization problem can be reliably solved using gradient descent based methods, as detailed in Algorithm \ref{alg:main}. We provide the complete proof of Theorem \ref{thm:h_convex} in the Appendix.

\begin{algorithm}[t]
\caption{HGPrompt: Training Process}
\label{alg:main}
\begin{algorithmic}[1]
\REQUIRE Target data $\mathcal{D}_T = \{(x_i, y_i)\}_{i=1}^N$, source prompts $\{P_j\}_{j=1}^M$, learning rate $\eta$, hyperparameter $\lambda$
\ENSURE Optimal weights $\boldsymbol{\alpha}^*$
\STATE Initialize $\boldsymbol{\alpha} = \{\alpha_1, \alpha_2, \ldots, \alpha_M\}$ with $\sum_{j=1}^M \alpha_j = 1$
\FOR{epoch = 1 to $K$}
    \STATE $\mu_y(\boldsymbol{\alpha}) = \mathbb{E}_{X|Y}[f_{\boldsymbol{\alpha}}(X)|Y=y]$
    \STATE $H(\boldsymbol{\alpha}) = \text{tr}(\text{cov}(f_{\boldsymbol{\alpha}})^{-1}\text{cov}(\{\mu_y\}))$
    \STATE Compute gradient variance: $\{v_i\}_{i=1}^M$ via Eq.(8)
    \STATE Evaluate gradient alignment regularization: $\mathcal{L}_{\text{align}}(\boldsymbol{\alpha})$ via Eq.(9)
    \STATE Compute total loss: $\mathcal{L}(\boldsymbol{\alpha}) = -H(\boldsymbol{\alpha}) + \lambda\mathcal{L}_{\text{align}}(\boldsymbol{\alpha})$
    \STATE Update weights: $\boldsymbol{\alpha} \leftarrow \boldsymbol{\alpha} - \eta\nabla_{\boldsymbol{\alpha}}\mathcal{L}$
\ENDFOR
\end{algorithmic}
\end{algorithm}

\subsection{Gradient Alignment Regularization}

Each prompt encodes task-specific knowledge. However, directly aggregating these prompts often leads to cross-interference between prompts, where independent evaluation fails to account for their synergistic or conflicting interactions. To address these issues, we propose aligning the gradient directions of all prompts, ensuring they collectively guide the model toward a unified optimization trajectory.

Building upon the gradient agreement principles from multi-task learning \cite{yu2020gradient, shi2021gradient, liu2023enhancing, rame2022fishr}, we propose a novel gradient variance matching objective for multi-source prompt transfer. Given $M$ source tasks with optimized prompts $\{P_i\}_{i=1}^M$, we compute the gradient of the loss with respect to each source prompt $P_i$ as:
\begin{equation}
g^i = \nabla_{P_i} \mathcal{L}(f_\theta([x_0;P_i; E(X)]), y).
\end{equation}
For each source prompt $P_i$, we compute its gradient variance:
\begin{equation}
v_i = \text{Var}(G) = \frac{1}{N-1} \sum_{j=1}^{N} (g_j^i - g^\alpha_j)^2,
\end{equation}
where $g^\alpha = \frac{1}{M}\sum_{i=1}^M \alpha_i g^i$ is the weighted mean of gradients, and $\mathbf{G} = [g^i]_{i=1}^M$ is the $N \times |P|$ prompt gradient matrix and $N$ is the batch size of samples used for gradient computation. We adapt the regularization to promote gradient variance alignment among source prompts:
\begin{equation}
\mathcal{L}_{\text{align}}(\boldsymbol{\alpha}) = \frac{1}{M}\sum_{i=1}^M \| v_i - \bar{v}(\boldsymbol{\alpha}) \|_2^2,
\end{equation}
where the mean gradient variance is defined as
$\bar{v}(\boldsymbol{\alpha}) =  \frac{1}{M}\sum_{i=1}^M v_i(\boldsymbol{\alpha})$. Balanced with a hyperparameter coefficient \(\lambda > 0\), this regularization penalty complements the original H-score objective,
\begin{equation}
\mathcal{L}(\boldsymbol{\alpha}) = -H(\boldsymbol{\alpha}) + \lambda \mathcal{L}_{\text{align}}(\boldsymbol{\alpha}).
\label{eq:total_loss}
\end{equation}

\subsubsection{Theoretical Analysis}
Our gradient alignment regularization \(\mathcal{L}_{\text{align}}\) builds on established theoretical foundations in domain-invariant learning \cite{parascandolo2020learning},  which seeks to identify invariant mechanisms in data by finding model parameters that exhibit consistent behavior across different domains. To quantify the consistency of the loss landscape around the optimal parameter \(\theta^*\) across domains, the inconsistency score is defined as follows:
\begin{figure}[t!]
  \centering
  \includegraphics[width=0.47\textwidth]{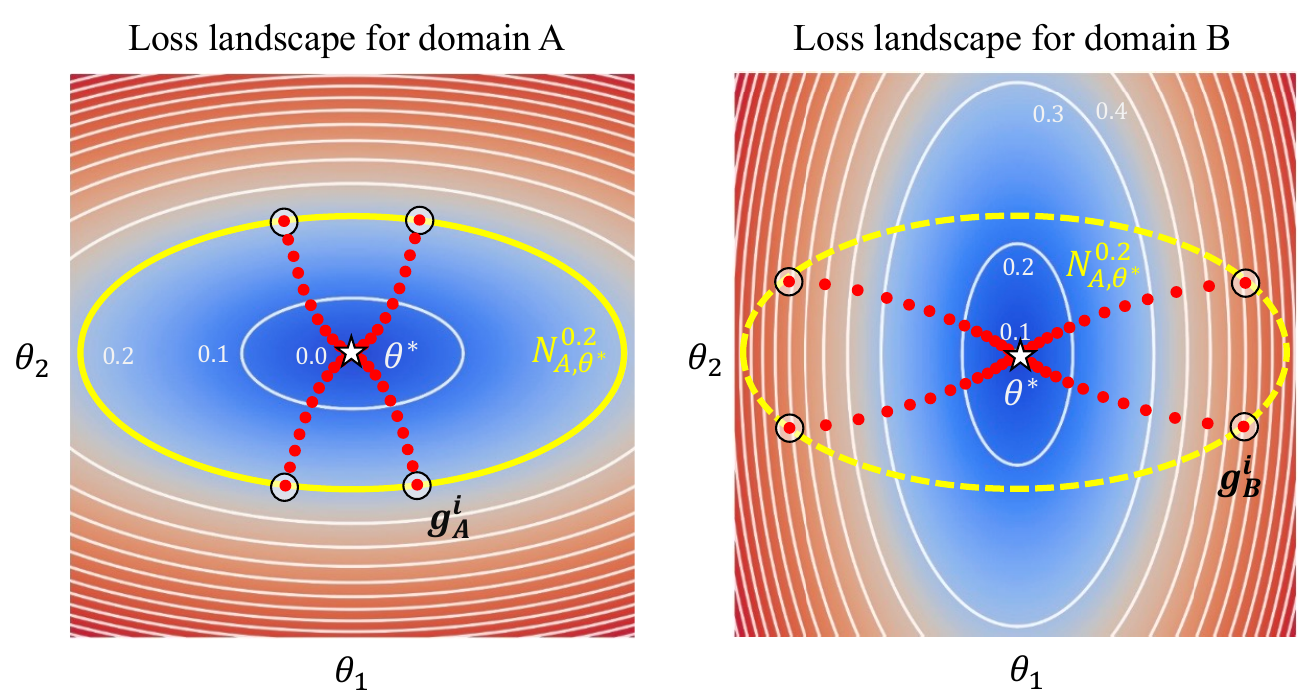} 
  \caption{Loss landscapes for a two-parameter model, showing conflicting gradient variances $\{g_i^{(A)}\}_{i=1}^{n_A}$ and $\{g_i^{(B)}\}_{i=1}^{n_B}$ around $\theta^*$. This shows the case where a nearby solution $\theta \in N_{A,\theta^*}^{\epsilon}$ maintains equivalent risk $\mathcal{R}_A(\theta) \approx \mathcal{R}_A(\theta^*)$ in domain A but exhibits higher risk in domain B.}
  \label{fig:fisher}
  \vspace{-0.3cm}
\end{figure}
\begin{definition}
Given a model parameter \(\theta^*\), the inconsistency score \(\mathcal{I}^{\epsilon}(\theta^*)\) is defined as:
\begin{equation}
    \mathcal{I}^{\epsilon}(\theta^*) = \max_{A,B} \; \max_{\theta \in N_{A,\theta^*}^{\epsilon}} \left|\mathcal{R}_B(\theta) - \mathcal{R}_A(\theta^*)\right|,
\end{equation}
where \(\theta \in N_{A,\theta^*}^{\epsilon}\) if there exists a continuous path in parameter space between \(\theta\) and \(\theta^*\) along which the risk \(\mathcal{R}_A\) remains within \(\epsilon\) of \(\mathcal{R}_A(\theta^*)\), for \(\epsilon > 0\).
\end{definition}
This concept is illustrated in Figure \ref{fig:fisher}, which demonstrates that minima with low consistency fail to generalize to new environments \cite{deutsch2011beginning}. The inconsistency score \(\mathcal{I}\) increases when the loss landscapes around $\theta^*$ present conflicting geometric structures across different domains.

\begin{theorem}
\label{the：appro}
Let $\theta^*$ be a simultaneous local minimum across domains with positive definite Hessians. Under the quadratic bowl assumption and for sufficiently small $\epsilon > 0$:
\begin{equation}
\mathcal{I}(\theta^*) = \max_{A,B} \bigg( |R_B(\theta^*) - R_A(\theta^*)| 
+ \max_{\frac{1}{2}\theta^\top H_A \theta \leq \epsilon} \frac{1}{2}|\theta^\top H_B \theta| \bigg).
\end{equation}
\end{theorem}
We provide the complete proof of Theorem \ref{the：appro} in the Appendix. The first term captures the loss landscape mismatch through domain-level risk differences. We will prove and show that $\mathcal{L}_{\text{align}}$ forces this term to be small in Appendix. For the second term, we employ a diagonal approximation of the Hessians for analysis. In that case, $H_e = \text{diag}(\lambda^e_1, \dots, \lambda^e_h)$ with $\forall i \in \{1,\dots,h\}, \lambda^e_i > 0$, the curvature term can be expressed as:
\begin{equation}
\begin{aligned}
\max_{\frac{1}{2}\theta^\top H_A\theta \leq \epsilon} \frac{1}{2}\theta^\top H_B\theta 
= \max_{\|\tilde{\theta}\|_2^2 \leq 2\epsilon} \sum_{i} \tilde{\theta}_i^2 \lambda^B_i / \lambda^A_i  \\
= \epsilon \cdot \max_i \lambda^B_i / \lambda^A_i.
\end{aligned}
\end{equation}
This result demonstrates that the second term diminishes when $H_A$ and $H_B$ have similar eigenvalues. Consequently, enforcing $H_A = H_B$ reduces inconsistencies in the loss landscape, thereby enhancing generalization performance. As we elaborate in the Appendix, our proposed $\mathcal{L}_{\text{align}}$ effectively aligns domain-level Hessians through gradient variance matching, by leveraging the fundamental connections between gradient variance, Fisher Information, and the Hessian.

\section{Experiments}

\begin{table*}[t]
\centering

\renewcommand{\arraystretch}{1.1}
\setlength{\tabcolsep}{5pt}
\newlength{\methodwidth}
\setlength{\methodwidth}{2cm} 

\begin{tabular}{m{\methodwidth}*{13}{c}|c}
\toprule[1pt]
\multicolumn{1}{l}{\textbf{Method}} & 
\rotatebox[origin=c]{90}{Cifar100} & 
\rotatebox[origin=c]{90}{DTD} & 
\rotatebox[origin=c]{90}{Flowers102} & 
\rotatebox[origin=c]{90}{Pets} & 
\rotatebox[origin=c]{90}{SVHN} & 
\rotatebox[origin=c]{90}{EuroSAT} & 
\rotatebox[origin=c]{90}{DMLab} & 
\rotatebox[origin=c]{90}{sNORB-Azim} & 
\rotatebox[origin=c]{90}{sNORB-Ele} & 
\rotatebox[origin=c]{90}{dSpr-Loc} & 
\rotatebox[origin=c]{90}{dSpr-Ori} & 
\rotatebox[origin=c]{90}{Clevr-Count} & 
\rotatebox[origin=c]{90}{Clevr-Dist} &
\rotatebox[origin=c]{90}{\textbf{Average}}
\\
\midrule
Linear & 61.7 & 58.6 & 96.6 & 83.9 & 32.7 & 83.9 & 30.6 & 12.2 & 20.3 & 12.6 & 18.2 & 32.1 & 28.6 & 44.0 \\
PARTIAL-1 & 64.4 & 60.3 & 97.5 & 86.0 & 36.3 & 87.8 & 32.5 & 16.5 & 21.8 & 31.3 & 39.2 & 41.3 & 32.1 & 49.8 \\
MLP-2 & 39.3 & 43.0 & 88.5 & 76.3 & 28.0 & 80.4 & 29.7 & 12.5 & 20.3 & 24.5 & 30.8 & 31.5 & 29.5 & 41.1 \\
MLP-3 & 41.9 & 46.2 & 90.5 & 78.4 & 30.3 & 83.9 & 30.7 & 14.1 & 21.5 & 25.9 & 33.1 & 33.8 & 30.2 & 43.1 \\
MLP-5 & 38.1 & 44.1 & 90.8 & 79.1 & 28.8 & 81.2 & 30.5 & 13.9 & 20.4 & 22.5 & 33.2 & 33.0 & 29.1 & 41.9\\
MLP-9 & 38.6 & 46.1 & 92.1 & 81.2 & 28.0 & 84.2 & 31.0 & 14.7 & 22.9 & 19.7 & 33.2 & 39.0 & 28.3 & 43.0 \\
\hline
Adapter & 73.8 & 61.7 & 97.5 & 86.6 & 32.7 & 85.3 & 29.4 & 11.9 & 19.5 & 22.4 & 20.8 & 40.1 & 35.1 & 47.4 \\
SIDETUNE & 53.5 & 58.7 & 93.4 & 77.2 & 17.6 & 37.2 & 26.7 & 10.6 & 15.1 & 13.2 & 13.6 & 20.3 & 19.4 & 35.1 \\
BIAS & 70.8 & 57.5 & 97.2 & 85.1 & 45.3 & 89.7 & 31.2 & 13.5 & 23.2 & 63.3 & \underline{39.7} & 49.1 & 54.5 & 56.2\\
\hline
VPT & 56.0 & 57.4 & 97.3 & 82.5 & 61.4 & 88.9 & 36.7 & 15.3 & 14.1 & 42.8 & 35.5 & 34.8 & 51.0 &51.8   \\
Single-Best & 63.9 & 60.2 & 97.0 & 83.4 & 63.2 & 89.5 & 36.1 & 18.3 & 18.9 & 57.1 & 36.4 & 38.3 & 51.9 & 54.9 \\
Average & 64.8 & 61.8 & 96.1 & 84.2 & 64.4 & 90.6 & 36.3 & 17.2 & 21.1 & 59.5 & 34.1 & 37.5 & 50.8 & 55.2 \\
SPoT & \underline{75.6} & \underline{63.7} & \underline{97.7} & \underline{86.3} & 70.4 & \underline{92.1} & 37.3 & \underline{19.4} & 23.3 & 65.0 & 36.0 & 41.5 & 52.8 & 58.5\\
ATTEMPT & 67.8 & 62.1 & 96.1 & 85.1 & 69.0 & 91.0 & 36.2 & 17.9 & 23.5 & 61.2 & 35.0 & \underline{43.5} & 51.2 & 56.9 \\
PANDA & 74.1 & 61.3 & 96.5 & 86.2 & \textbf{71.2} & 90.8 & \underline{37.8} & \underline{19.4} & \underline{24.0} & \underline{67.7} & 37.3 & 42.8 & \textbf{53.9} & \underline{58.7} \\
\hline
\rowcolor{gray!30} HGPrompt & \textbf{75.9} & \textbf{64.2} & \textbf{98.1} & \textbf{87.4} & \underline{71.0} & \textbf{92.6} & \textbf{38.1} & \textbf{20.3} & \textbf{24.9} & \textbf{68.1} & \textbf{40.4} & \textbf{49.3} & \underline{53.5} & \textbf{60.3} \\
\bottomrule[1pt]
\end{tabular}
\caption{Performance comparison across diverse vision tasks using a Vision Transformer (ViT-B/16) backbone pre-trained on ImageNet-21k. The second-best results are underlined, while the best results are highlighted in bold. All reported values represent the average accuracy obtained from three independent runs, with the highest average accuracy achieved by our method.}
\label{tab:comparisons}
\end{table*}

We evaluate the proposed approach for a wide range of downstream recognition tasks with pre-trained Transformer backbones. We first describe our experimental setup, including the pre-trained backbone and downstream tasks and a brief introduction to other transfer learning methods.
\subsection{Setup}\label{expsetup}
\subsubsec{Datasets} 
We experiment on a collection of 13 datasets from V-tab-1k \cite{zhai2019large}. VTAB is a collection of dieverse visual  classification tasks, which encompasses three distinct categories of tasks: Natural, featuring images taken with conventional cameras; Specialized, containing data acquired through specialized devices, such as satellite sensors; and Structured, which demands spatial reasoning, like counting objects. We provide more detailed descriptions of the datasets in the Appendix.

\subsubsec{Implementation Details}We implement all experiments on NVIDIA A800-80GB GPUs. For a fair comparison, all methods use a ViT-B/16 backbone pre-trained on ImageNet-21k, and the number of prompt tokens is set to 50. We follow the original configurations, eg. number of image patches divided, existence of \texttt{[CLS]}, etc. We train the prompt on all the source tasks for 10 epochs for source prompt training. We use 2000 samples from each source task for each target task to compute the transferability loss and gradient alignment loss. 

\subsubsec{Baselines}
We compare our approach to eleven recent methods, categorizing them as follows: (1) Methods that retrain the classification head: \textbf{PARTIAL-$k$} \cite{partialk2016} fine-tunes only the last $k$ layers of the backbone while freezing others; \textbf{MLP-$k$} utilizes a multilayer perceptron with $k$ layers as the classification head instead of a linear layer. (2) Methods that update a subset of backbone parameters or add new trainable modules: \textbf{Adapter} \cite{pmlr-v97-houlsby19a} inserts new MLP modules with residual connections into transformer layers; \textbf{SIDETUNE} \cite{zhang2020side} trains a side network and linearly interpolates between pre-trained features and side-tuned features before feeding them into the head; \textbf{BIAS} \cite{ben-zaken-etal-2022-bitfit} fine-tunes only the bias terms of the pre-trained backbone. (3) Prompt transfer methods: \textbf{Average} directly uses the mean of source prompt embeddings; \textbf{Single-Best} selects the source prompt with optimal transfer performance; \textbf{Visual Prompt Tuning (VPT)} \cite{jia2022visual} initializes target prompt embeddings randomly; \textbf{SPoT} \cite{vu-etal-2022-spot} calculates similarity between source and target prompt embeddings; \textbf{ATTEMPT} \cite{asai-etal-2022-attempt} mixes pre-trained source and target prompts via an attention mechanism; \textbf{PANDA} \cite{panda-2024} measures cosine similarity between task embeddings as a transferability proxy.

\begin{figure}[t]
    \centering
    \includegraphics[width=0.99\linewidth]{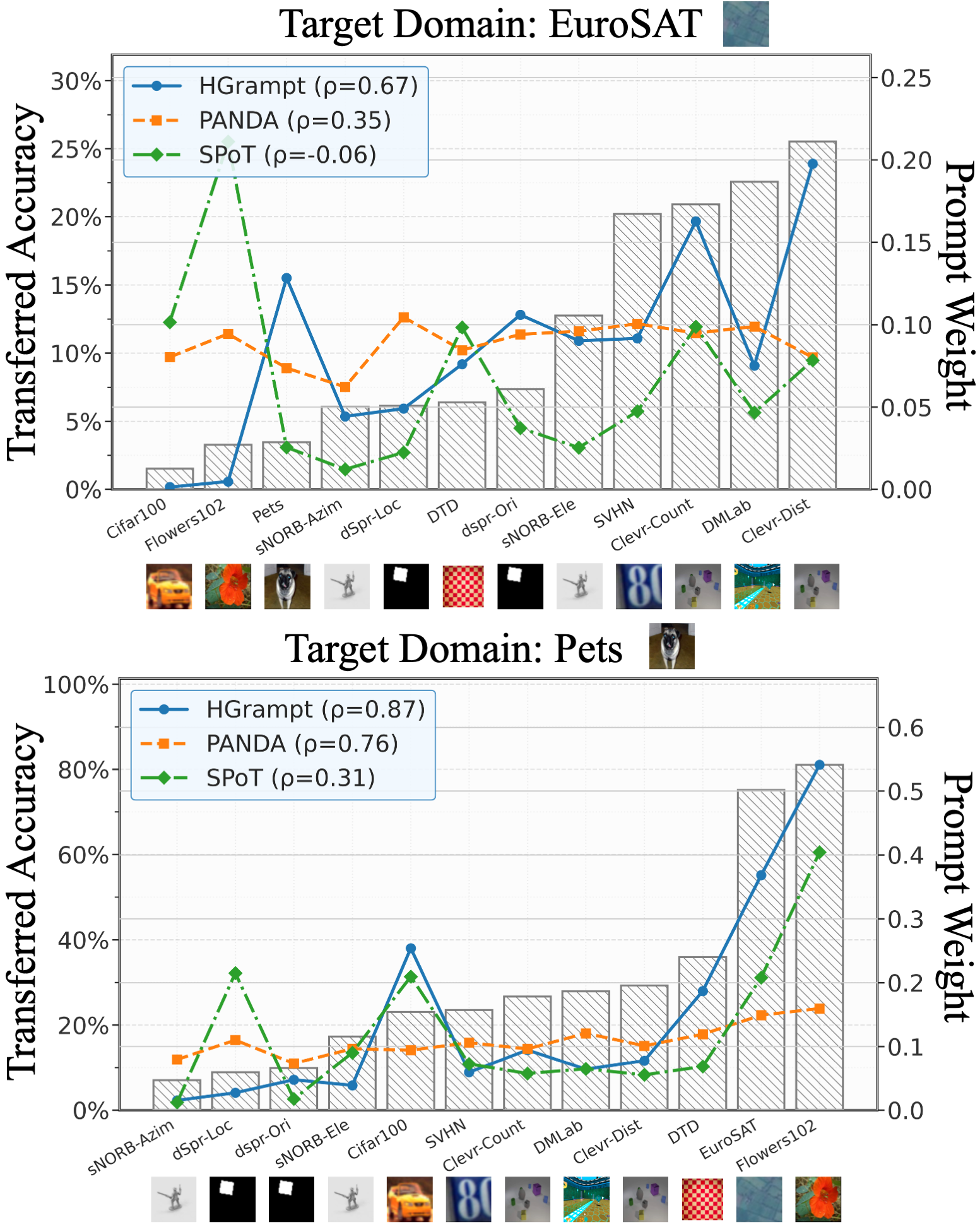}
    \caption{Prompt Weights analysis for 12 source prompts. Bar plots represent single-source transfer accuracy (left axis), while line plots indicate prompt weights (right axis).}
    \label{fig:corr}
    \vspace{-0.3cm}
\end{figure}



\subsection{Main Results}
Our experimental evaluation across 13 diverse vision tasks, as detailed in Tab.~\ref{tab:comparisons}, demonstrates HGPrompt's superiority over 13 baselines using a ViT-B/16 backbone pre-trained on ImageNet-21k. The proposed method achieves state-of-the-art performance with an average accuracy of 60.3\%, surpassing prior multi-source prompt transfer approaches.
HGPrompt excels in fine-grained recognition tasks, achieving top results on Flowers102 and Oxford Pets. It also outperforms all baselines in texture analysis on DTD and maintains competitive performance on CIFAR100. Notably, the method establishes new state-of-the-art results in geometric reasoning tasks, including sNORB-Azimuth and dSprite-Orientation, with significant improvements in complex visual reasoning tasks like Clevr-Count. While PANDA retains an advantage in SVHN, HGPrompt exhibits a more balanced and robust performance across all task categories, highlighting its effectiveness.

\begin{table}[t]
\centering
\caption{Ablation Study on Framework Components}
\label{tab:ablation}
\begin{tabular}{cc|ccccc}
\toprule
$\mathbf{H(\alpha)}$ & $\mathcal{L}_{\text{align}}$ & Cifar & DTD & Pets & Euro &\textbf{Avg} \\
\midrule
$\times$ & $\times$ & 60.4 & 57.8 & 82.7 & 89.1 & 72.5 \\
$\checkmark$ & $\times$ & 74.6 & 62.3 & 85.9 & 91.2 & 78.5 \\
$\times$ & $\checkmark$ & 74.1 & 61.9 & 85.5 & 90.8 & 78.1 \\
$\checkmark$ & $\checkmark$ & \textbf{75.9} & \textbf{64.2} & \textbf{87.4} & \textbf{92.6} & \textbf{80.0} \\
\bottomrule
\end{tabular}
\end{table}
\subsection{Ablation Study}
The ablation study in Table~\ref{tab:ablation} systematically evaluates each component's contribution in our framework. Additional dataset results are provided in the Appendix. The baseline method, which directly optimizes weights by minimizing cross-entropy loss on the target task, achieves 72.5\% average accuracy. Using the H-score objective alone improves performance to 78.5\%, validating its effectiveness for feature discriminability evaluation. Similarly, employing $\mathcal{L}_{\text{align}}$ as the sole optimization objective yields 78.1\% accuracy. Most notably, combining both components produces the best performance 80.2\%, demonstrating their complementary roles in achieving optimal transfer learning results.

\subsection{Analysis and Discussion}
\subsubsection{Evaluation on Prompt Weights}
To demonstrate the effectiveness of our learnt Prompt Weights, we pretrained a set of source prompts and evaluated their zero-shot transfer accuracy, as shown in Fig.~\ref{fig:corr}. We plotted the weights calculated by SPoT, PANDA, and our proposed HGPrompt method. Our approach more accurately reflects semantic task affinities, indicating that our proposed metric can better distinguish different task relationships: similar tasks exhibit larger prompt transferability. For example, tasks involving natural scenes—such as Flowers, Pets, and DTD—demonstrate higher inter-task transferability, a pattern largely captured by HGPrompt. To systematically validate this finding, we conducted a quantitative analysis presented in Table~\ref{tab:spearman_results}. We computed Spearman's rank correlation between the predicted weights and the actual zero-shot transfer accuracy, confirming that our metric achieves superior correlation compared to existing approaches. In contrast, the results reveal that SPoT and PANDA struggle to accurately evaluate task-relevant semantic information, exhibiting significant fluctuations. Complete results are provided in the appendix.

\begin{figure}[t]
    \centering
    \includegraphics[width=0.9\linewidth]{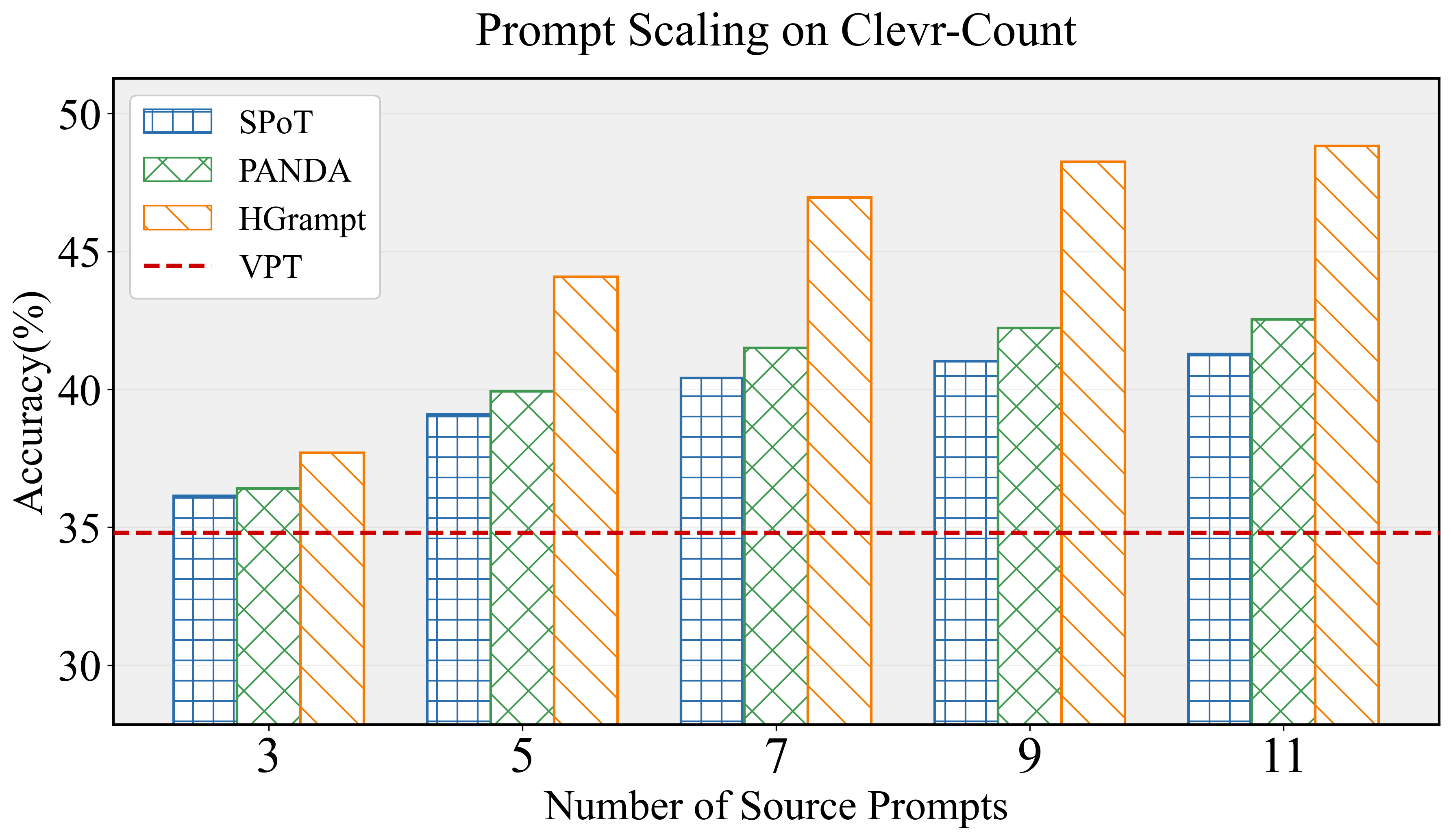}
    \caption{Performance scaling with increasing source prompts on Clever-Count target domain.}
    \label{fig:scaling}
    \vspace{-0.3cm}
\end{figure}

\begin{table}[t]
\centering
\setlength{\tabcolsep}{4pt}
\renewcommand{\arraystretch}{0.8}
\caption{Spearman's $\rho$ correlation scores.}
\label{tab:spearman_results}
\begin{tabular}{lccccc|c}
\toprule
 & Cifar & C-di & d-Lo & DML & SVHN & \textbf{Avg}  \\
\midrule
SPoT     & 0.552 & 0.175 & -0.168 & 0.112 & -0.147 & 0.105 \\
PANDA    & 0.916 & 0.441 & 0.552 & 0.713 & 0.224 & 0.569 \\
\rowcolor{gray!30} HGPrompt & \textbf{0.944} & \textbf{0.664} & \textbf{0.853} & \textbf{0.727} & \textbf{0.853} &  \textbf{0.808} \\
\bottomrule
\end{tabular}
\end{table}


\begin{figure}[t]
    \centering
    \begin{subfigure}[b]{0.11\textwidth}  
        \includegraphics[width=\textwidth]{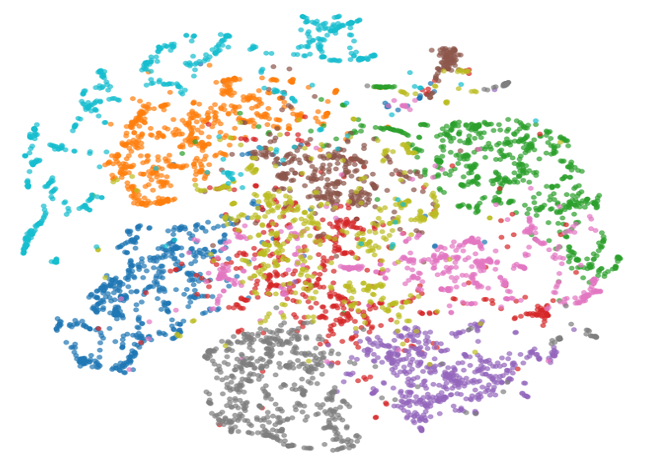}
        \caption{SPoT}
    \end{subfigure}
    \hspace{0.001\textwidth} 
    \begin{subfigure}[b]{0.11\textwidth}
        \includegraphics[width=\textwidth]{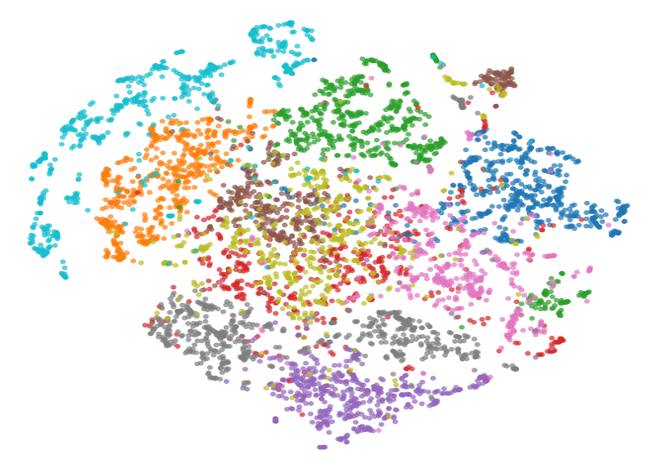}
        \caption{ATTEMPT}
    \end{subfigure}
    \hspace{0.001\textwidth}
    \begin{subfigure}[b]{0.11\textwidth}
        \includegraphics[width=\textwidth]{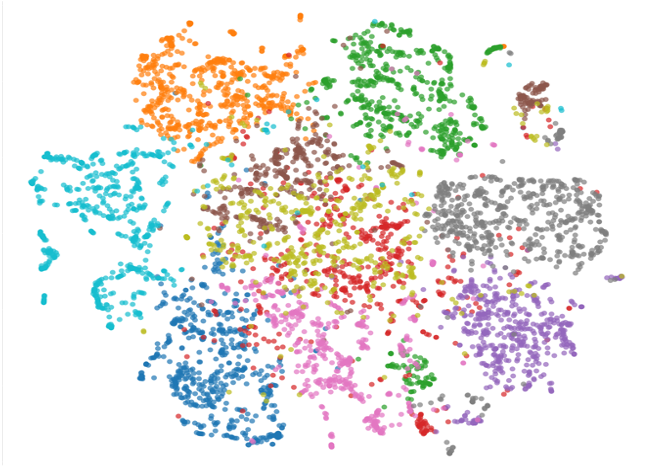}
        \caption{PANDA}
    \end{subfigure}
    \hspace{0.001\textwidth}
    \begin{subfigure}[b]{0.11\textwidth}
        \includegraphics[width=\textwidth]{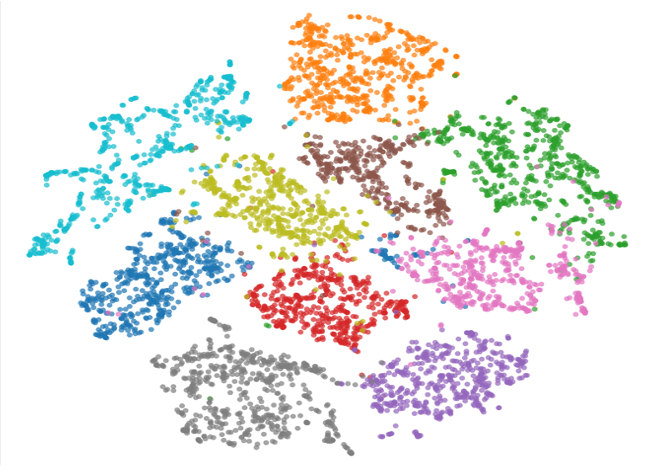}
        \caption{HGPrompt}
    \end{subfigure}
    \caption{t-SNE Visualization of representations on EuroSAT (10 Classes). Each color corresponds to a distinct class.}
    \label{fig:t-sne}
\end{figure}
\subsubsection{Performance Scaling with Source Prompts Number} As shown in Fig.5, our method demonstrates progressively stronger performance advantages over PANDA and SPOT when using DTD as the target domain, particularly as the number of source prompts increases from 3 to 11. Results for other domains can be found in Appendix. This scaling behavior highlights our approach's superior capability in effectively utilizing larger prompt collections. While absolute accuracy shows consistent improvement with additional source prompts, the system eventually approaches an inherent efficiency ceiling.

\subsubsection{Representation Space Visualization} To analyze the effect of prompt ensemble on the learned representation of the target test data, we present t-SNE visualizations of ViT feature embeddings in four different prompt transfer methpds in Fig.\ref{fig:t-sne}. 
As shown in Fig.\ref{fig:t-sne}.d, our method shows better class discriminability than other baselines. Instead of scattered clusters, objects from the same category form tightly grouped regions with clear separation boundaries. 
The visualization underscores the effectiveness of our method in constructing a coherent and well-structured feature space for transfer learning.

\subsubsection{Parameter Analysis}
We analyze how the regularization coefficient $\lambda$ influences our results. Performance trends across benchmarks reveal optimal outcomes when both objectives contribute comparably, indicating mutual reinforcement between feature transferability and gradient alignment. The full results are provided in the appendix.  

\subsubsection{Discussion}
While our current work has demonstrated the effectiveness of visual prompting within transformer architectures, we acknowledge its limitations in terms of architectural specificity and modality constraints. Future work in these directions may require novel approaches to prompt design and adaptation, potentially drawing inspiration from recent advances in multimodal learning and architecture-agnostic representation techniques. The ultimate goal would be to establish prompting as a truly universal interface for model adaptation and control, transcending specific architectural choices or modality limitations.

\section{Conclusion}
In this work, we introduce HGPrompt, a novel framework for multi-source prompt transfer that explicitly optimizes the prompt ensemble. Our methodology determines optimal source weights by maximizing the H-score while matching gradient variance, thereby effectively quantifying the transferability of the source prompt ensemble. By dynamically balancing feature discriminability with generalization, HGPrompt leverages complementary information across prompts while simultaneously suppressing interference. Our contributions establish a solid foundation for advancing multi-source prompt transfer, offering both theoretical and practical insights to enhance foundation model adaptability.

\bibliography{main}

\end{document}